\ifdefined\mainsupp
\else

\documentclass[runningheads]{llncs}
\usepackage[T1]{fontenc}
\usepackage{graphicx}

\usepackage[colorlinks,urlcolor=blue,linkcolor=blue,citecolor=blue]{hyperref}

\usepackage{amssymb}            
\usepackage{mathrsfs}           
\usepackage{amsmath}
\usepackage{mathtools}          

\mathtoolsset{showonlyrefs}     
\usepackage{subcaption}         
\usepackage[space]{grffile}     
\usepackage{url}                

\usepackage{times}
\usepackage{soul}
\usepackage{url}
\usepackage{booktabs}
\usepackage{algorithm}
\usepackage[switch]{lineno}

\usepackage{amsfonts}       
\usepackage{nicefrac}       
\usepackage{microtype}      
\usepackage{xcolor}         
\usepackage{algpseudocode}

\usepackage{float}
\usepackage{flafter}
\usepackage{placeins}
\usepackage{xr}
\newcommand{\DV}{$DV3'$}

\newboolean{anon}
\setboolean{anon}{false} 

\begin{document}

\fi

\title{Augmenting Replay in World Models for Continual Reinforcement Learning}

\ifthenelse{\boolean{anon}}
{
\author{First Author\inst{1}\orcidID{0000-1111-2222-3333} \and
Second Author\inst{2,3}\orcidID{1111-2222-3333-4444} \and
Third Author\inst{3}\orcidID{2222--3333-4444-5555}}
\authorrunning{F. Author et al.}
%
\institute{Institute 1
\email{email@email.com} \and
Institute 2 
\email{\{abc,lncs\}@email.com}
\url{http://www.institute.com} }
}
{
\author{Luke Yang\inst{1} \and
	Levin Kuhlmann\inst{1}\orcidID{0000-0002-5108-6348} \and
	Gideon Kowadlo\inst{1,2}\orcidID{0000-0001-6036-1180} }

\authorrunning{L. Yang et al.}

\institute{Monash University
\email{levin.kuhlmann@monash.edu} \\
\and
Cerenaut
\email{gideon@cerenaut.ai} \\
\url{https://cerenaut.ai}}

}

\maketitle

\begin{abstract}
Continual RL requires an agent to learn new tasks without forgetting previous ones, while improving on both past and future tasks. The most common approaches use model-free algorithms and replay buffers can help to mitigate catastrophic forgetting, but often struggle with scalability due to large memory requirements. Biologically inspired replay suggests replay to a world model, aligning with model-based RL; as opposed to the common setting of replay in model-free algorithms. Model-based RL offers benefits for continual RL by leveraging knowledge of the environment, independent of policy. We introduce WMAR (World Models with Augmented Replay), a model-based RL algorithm with a memory-efficient distribution-matching replay buffer. WMAR extends the well known DreamerV3 algorithm, which employs a simple FIFO buffer and was not tested in continual RL. We evaluated WMAR and DreamerV3, with the same-size replay buffers. They were tested on two scenarios: tasks with shared structure using OpenAI Procgen and tasks without shared structure using the Atari benchmark. WMAR demonstrated favourable properties for continual RL considering metrics for forgetting as well as skill transfer on past and future tasks. Compared to DreamerV3, WMAR showed slight benefits in tasks with shared structure and substantially better forgetting characteristics on tasks without shared structure. Our results suggest that model-based RL with a memory-efficient replay buffer can be an effective approach to continual RL, justifying further research.


\end{abstract}

\section{Introduction}

Typically, RL focuses on a single unchanging task.
Continual RL, however, presents tasks sequentially,  representing many real-world scenarios \cite{khetarpal_towards_2022}.
Continual RL poses the well-known challenge of catastrophic forgetting \cite{mccloskey_catastrophic_1989}, where an agent forgets old tasks when learning new ones. Another challenge is to use prior knowledge to learn new tasks more efficiently.
%
%
One approach to continual RL is to store experiences of all tasks to prevent catastrophic forgetting \cite{lipton_combating_2018}. This is often done through a replay buffer.
 However, it requires very high storage capacity, reducing scalability and accessibility of the algorithm \cite{openai_dota_2019}.

The inspiration for many replay-based methods comes from Complementary Learning Systems (CLS) \cite{hassabis_neuroscience-inspired_2017,khetarpal_towards_2022}, which describes learning in mammalian brains. The hippocampus memorises recent observations and replays them to the neocortex, which is a slow statistical learner. Replay is interleaved with new experiences, thus mitigating catastrophic forgetting. 
Interestingly, the neocortex is understood to form a world model whose purpose is to predict the consequences of our actions \cite{mathis_neocortical_2023}. 
Although CLS describes replay to a world model, traditionally replay in RL is used in model-free RL to improve the policy, rather than a world model directly.

The idea of world models has been exploited in model-based RL algorithms, where a world model is capable of predicting the effects of actions on the environment \cite{ha_world_2018,hafner_learning_2019} and the more recent state-of-the-art DreamerV1 to V3 \cite{hafner_dream_2020,hafner_mastering_2022,hafner_mastering_2023}. They have been applied to continual RL by several authors 
\cite{nagabandi2018deep,huang2021continual,kessler_effectiveness_2023,rahimi-kalahroudi_replay_2023}.
World models are an intuitive choice for exploiting replay buffers, as they naturally support off-policy learning.

In this paper, we take DreamerV3 \cite{hafner_mastering_2023}, add a memory-efficient replay buffer \cite{isele_selective_2018}, and apply it to continual RL.
Hence, we present World Models with Augmented Replay (WMAR).
%
We applied WMAR to two settings. 
First, where each task has a distinct environment and reward function; the most common setting used in continual RL, and Atari games are often used.
In the second setting, there is commonality between tasks, and learnt knowledge can be leveraged to perform subsequent tasks. This is referred to as tasks with `shared structure' \cite{khetarpal_towards_2022}. Often a video game is used, but conditions such as movement dynamics or spacing of features in the environment change over time, e.g. \cite{riemer_learning_2019}. Many potential real-world applications exist within the `shared structure' setting. For example, a robot assistant that should be able to acquire new and related tasks within a home, like cleaning with a broom and then a mop.
We used OpenAI Procgen (procedurally generated games) for tasks with `shared structure', and Atari games for tasks `without shared structure'. Our analysis is not restricted to catastrophic forgetting but included backward and forward transfer of skills between tasks. We aim to achieve the following qualities of a successful CL agent \cite{khetarpal_towards_2022,chen_continual_2018}:

\begin{itemize}
    \item \textbf{Stability} avoid forgetting and losing performance on previously learnt tasks
    \item \textbf{Backward transfer} increase performance on previously learnt tasks after training on new, similar tasks.
    \item \textbf{Plasticity} don't slow down on learning new and dissimilar tasks compared to learning them independently
    \item \textbf{Forward transfer} learn new similar tasks faster compared to learning them independently
%
%
%
    \item \textbf{Scalable} low memory and computational requirements
    \item \textbf{Task id's} don't rely on task identifiers, which are often unavailable
\end{itemize}

The primary contributions include: a) applying model-based RL to continual RL for tasks with and without shared structure and b) augmenting the replay buffer of DreamerV3 with a memory-efficient long-term distribution matching buffer.
For background and related work, see S.\ref{app:background}.


\section{World models with augmented replay (WMAR)}

WMAR extends DreamerV3, which has achieved state-of-the-art performance in several single-GPU RL benchmarks. 
WMAR consists of three primary components. A \textbf{world model} for modelling the environment, \textbf{actor-critic controller} for acting on the environment, and \textbf{augmented replay buffer} to store past experiences.
The replay buffer is used to train the world model, and the world model is able to simulate `dreamed' experiences, which are used to train the controller.
A world model's ability to simulate the environment enables off-policy learning and data augmentation, which are beneficial in continual learning where direct environment interaction may be limited.
We hypothesise that maintaining the world model's accuracy across different tasks will help preserve performance on past environments while adapting to new ones. This approach does not require explicit task identifiers, potentially allowing for more flexible adaptation to changing environments.

The key components are described below, and more details are found in S.\ref{app:world_models}. 
The source code is available at 
\ifthenelse{\boolean{anon}}
{
	\url{https://anonymous.4open.science/r/WMAR-FB67}.
}
{
	\url{https://github.com/cerenaut/wmar}.
}

\subsection{World model}

\begin{figure}
  \centering
  \begin{subfigure}{0.49\textwidth}
	  \includegraphics[width=\linewidth]{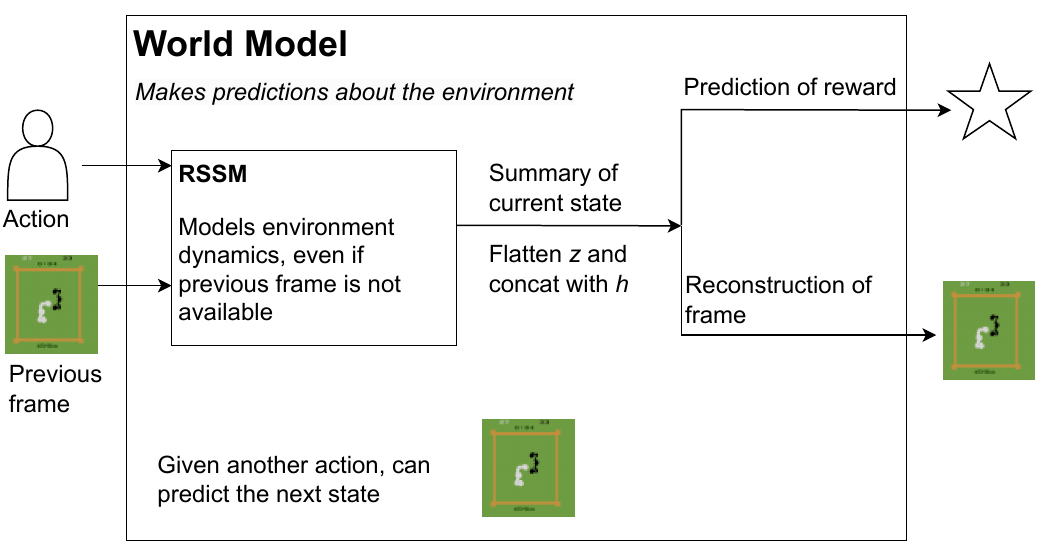}
	  \caption{World Model}	  
  \end{subfigure}%
  \begin{subfigure}{0.49\textwidth}
	  \includegraphics[width=\linewidth]{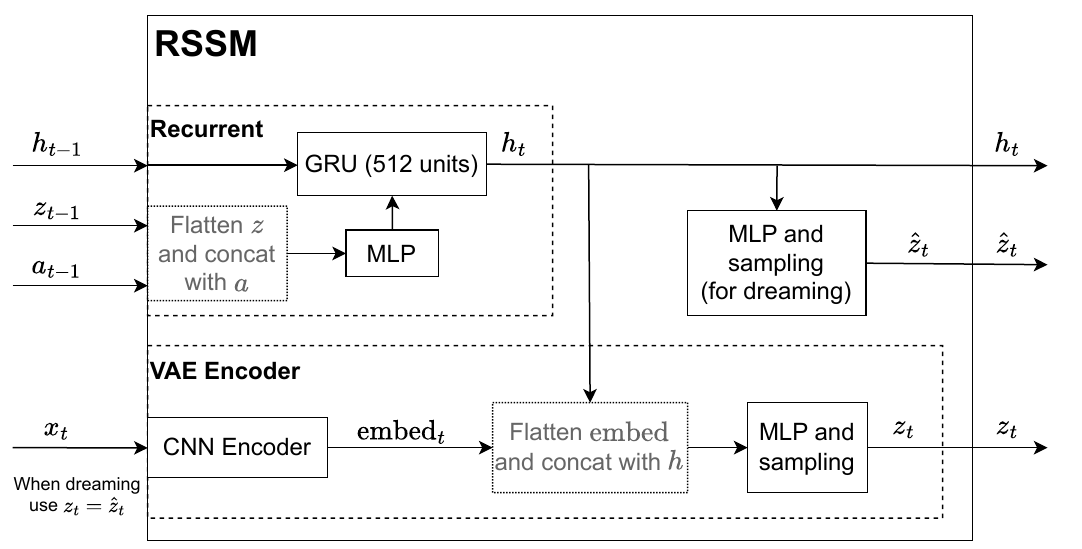}
	  \caption{Flow of data within the RSSM}
  \end{subfigure}
  \caption{World Model overview and RSSM details.}
  \label{fig:world_model}
\end{figure}

A Recurrent State-Space Model (RSSM) \cite{hafner_learning_2019}, predicts environment dynamics, Figure~\ref{fig:world_model}. 
It maintains a deterministic hidden state $h_t$ and models a stochastic representation of the next state given current and previous observations $x_{1:t}$ and actions $a_{1:t}$
Dynamics are modelled with a Gated Recurrent Unit (GRU) network, predicting the deterministic state $h_{t+1} = \text{GRU}(h_t, z_t, a_t)$ and the stochastic state $\hat{z}_{t+1} = f(h_{t+1})$. 
The stochastic state $z_t$ is either inferred by a variational autoencoder $z_t \sim q_{\theta}(z_t \mid h_t, x_t)$ or predicted by the dynamics model $\hat{z}_t \sim p_{\theta}(\hat{z}_t \mid h_t)$ for open-loop prediction in dreaming, where stochastic state posteriors $z_t$ are unavailable.



We used a standard GRU with Tanh activation for the recurrent component.
The stochastic state $z_t$ comprises 32 discrete stochastic units, each with 32 categorical classes, following the architecture of DreamerV3.


The world model state at timestep $t$ is the concatenation of $h_t$ and $z_t$, forming a Markovian representation of the environment state.
The world model is trained to reconstruct input images and rewards.
KL balancing \cite{hafner_mastering_2022} helps to model the transition between states.

\subsection{Actor critic}

The actor and critic (S.\ref{app:network_arch}) are MLPs that map the state of the model to actions and value estimates. They are trained entirely on trajectories generated stochastically by the world model, referred to as `dreaming'. 
The actor and critic are trained on-policy using REINFORCE \cite{williams_simple_1992}. Inevitable changes in the model state space are not an issue since imagined trajectories are cheap to generate and do not require interaction with the actual environment, allowing this process to be run to convergence.

\subsection{Augmented replay buffer}

\begin{figure}
  \centering
  \includegraphics[width=0.6\textwidth]{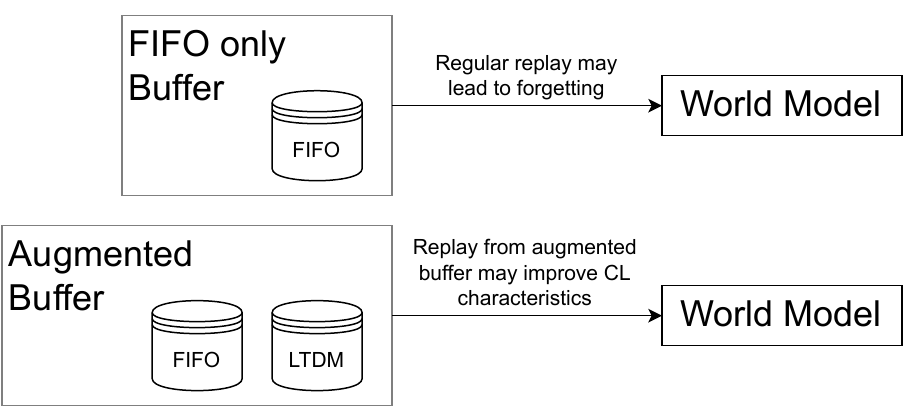}
  \caption{The use of replay within the world model context. LTDM refers to the long-term distribution matching buffer.}
  \label{fig:replay_buffers}
\end{figure}

The augmented replay buffer, \autoref{fig:replay_buffers}, comprises a short-term FIFO buffer $\mathcal{D}_1$ (as used in DreamerV3) and a long-term global distribution matching buffer $\mathcal{D}_2$. They are equally sized and used in parallel. 
Data from both are uniformly sampled for each training minibatch (S.Algorithm~\ref{alg:combined_buffers}).
A key objective is to minimise memory requirements and hence the size of the replay buffer. While Dreamer maintained a single buffer containing the last 1 million (1,000,000) observations, we empirically chose a size of $2^{18} \approx$ 262,000 observations for each buffer, resulting in an augmented buffer that is significantly smaller (about $2\times$), without noticeably affecting performance. We also introduced \textit{spliced rollouts}, which are a simple and sometimes necessary alternative to storing entire episodes, enabling smaller buffer sizes. 

\paragraph{Short-term FIFO buffer}
The FIFO buffer has a capacity of $2^{18}$ samples and contains the most recent rollout observations. 
The buffer allows the world model to train on all incoming experiences and biased the training samples to those more recently collected, thus improving convergence properties on the current task.

\paragraph{Long-term global distribution matching (LTDM) buffer}
Matching the global training distribution even with limited capacity in the replay buffer can reduce catastrophic forgetting \cite{isele_selective_2018}.
Therefore, we used a long-term global distribution matching buffer, also with a capacity of $2^{18}$ samples. 
It contains a uniform random subset of 512 spliced rollouts. 
Reservoir sampling was used, assigning a random value for each rollout chunk as a key in a size-limited priority queue, preserving experiences with the highest key values and discarding the remainder.

\paragraph{Spliced rollouts} Restricting the size of the replay buffer can cause it to become full with only a few episodes. Containing too few unique episodes may cause the training data to become highly unrepresentative of the general environment states, thus reducing the world model's accuracy and likely causing a subsequent loss of performance from the actor. This is especially relevant for the long-term global distribution matching buffer, which may store relatively few samples from each environment.
Therefore, while typical implementations store complete rollouts, we spliced rollouts into smaller chunks of size 512. Operating over the spliced rollouts rather than entire rollouts provided a guarantee on the granularity for sampling data.
The remaining rollouts with fewer than 512 states were concatenated before the next episode with an appropriate reset flag indicating the start of a new episode. To improve the efficiency of the rollout process, we truncated episodes after a fixed number of steps, so that the final number of environment steps was identical in every training iteration. 
We found that spliced rollouts were a simple solution to control granularity in the distribution matching buffer and did not adversely affect performance.

\subsection{Task-agnostic exploration}
In tasks without shared structure, there are significant differences between environments in terms of task dynamics, visual differences, and magnitude of rewards, which poses a significant challenge to CL, especially without task id's.
Exploring new environments is difficult, as policies that have already been trained on previous tasks may lack the randomness required to adequately explore the state-space of a new task.
We addressed this challenge using fixed-entropy regularisation (as in DreamerV3) for training the world model.
In addition, we used predetermined reward scales in individual environments based on single-task baselines, during actor-critic training.
This approach enabled mitigation of the exploration challenge without the addition of an exploration-orientated learning system such as Plan2Explore \cite{sekar_planning_2020}. 

\section{Experiments}

\subsection{Dataset}
We evaluated WMAR in a set of challenging OpenAI Procgen and Atari environments \cite{cobbe_leveraging_2020,bellemare_arcade_2013}.
They are commonly used RL benchmarks, can be extremely similar for the case of shared structure (Procgen) or dissimilar where there is no shared structure (Atari), and they were computationally feasible on the available hardware.
We define shared structure as tasks that have similar environment state and observations, action dynamics, and rewards. We applied visual perturbations to create differences between tasks. To evaluate performance on tasks without shared structure, we selected 4 Atari tasks.

\subsection{Baselines}
To measure the efficacy of the augmented replay buffer (i.e., adding a long-term distribution matching buffer with spliced rollouts), we compared WMAR and DreamerV3, with equal memory allowance for replay buffers. 
For the most direct comparison, we implemented our own version of DreamerV3, referred to as \DV, and augmented the replay buffer to create WMAR. 
To validate DV3', we compared to an open source implementation of DreamerV3 \url{https://github.com/danijar/dreamerv3}, see S.\ref{app:validation}.

We also ran single-task baselines: WMAR and a random agent.
The single-tasks baselines were used for evaluation metrics.

\subsection{Evaluation metrics}

We evaluated agents using task reward and following Kessler et al. \cite{kessler_effectiveness_2023}, extended the evaluation to forgetting and forward transfer.
Forgetting informs us about backward transfer and stability, and forward transfer informs us about plasticity.

We evaluated performance by normalising episodic reward to two single-task baselines: WMAR and a random agent.
The normalised comparison allowed us to measure the relative performance, providing a natural evaluation of forgetting and forward transfer for all environments in each CL suite.

\newcommand{\rf}{\text{ST}}
\subsubsection{Normalised rewards}

We define an ordered suite of tasks $\boldsymbol{\mathcal{T}} = (\tau_1, \tau_2, \dots, \tau_T)$. 
Performance in task $\tau \in \boldsymbol{\mathcal{T}}$ after $n$ steps in single-task experiments is given by $p_{\rf_\tau}(n)$ and in CL by $p_\tau(n)$.
Agents were trained on each task for $n=N$ environment steps in single-task and CL experiments. 
For each task $\tau \in \boldsymbol{\mathcal{T}}$, we calculated the normalised reward using the average episodic rewards of the single-task random policy $p_{\rf_\tau}(0)$ and the trained agent after $n$ environment steps $p_{\rf_\tau}(n)$.

\begin{align}
    q_\tau(n) = \frac{p_\tau(n) - p_{\rf_\tau}(0)}{p_{\rf_\tau}(n) - p_{\rf_\tau}(0)}\label{eq:normalization}
\end{align}

A normalised score of 0 corresponds to random performance and a score of 1 corresponds to the performance when trained on only that task (training on only one task is usually much easier than learning the same task in a continual learning setting).

\subsubsection{Forgetting (Backward transfer)}

Average forgetting for each task is the difference between performance after training on a given task and performance at the end of all tasks. Average forgetting over all tasks is defined as:

\begin{align}
    F = \frac{1}{T} \sum_{\tau=1}^{T} q_{\tau}(\tau \times N) - q_{\tau}(T \times N)
\end{align}

A lower value for forgetting is indicative of improved \textit{stability} and a better continual learning method. A negative value for forgetting would imply that the agent has managed to gain performance on earlier tasks, thus exhibiting \textit{backward transfer}.

\subsubsection{Forward transfer}

The forward transfer for a task is the normalised difference between performance in the CL and single-task experiments. The average over all tasks is defined as:

\begin{align}
    FT &= \frac{1}{T} \sum_{\tau=1}^{T} \frac{S_\tau - S_{\rf_\tau}}{S_{\rf_\tau}} \\
    \intertext{where}
    S_\tau &= \frac{1}{N} \sum_{n=1}^{N} q_{\tau}((\tau-1) \times N + n) \\
    S_{\rf_\tau} &= \frac{1}{N} \sum_{n=1}^{N} q_{\rf_\tau}(n)
\end{align}

The larger the forward transfer, the better the continual learning method. A positive value implies effective use of learnt knowledge from previous environments and, as a result, accelerated learning in the current environment. When each task is not related to the others, no positive forward transfer is expected. In this case, forward transfer of 0 represents optimal \textit{plasticity}, and negative values indicate a barrier to learning newer tasks from previous tasks.

\subsection{Experimental setup}
All WMAR experiments are trained on one Nvidia A40 40GB GPU with all single-task benchmarks running within 0.25 days and continual learning benchmarks running within 1 day of wall time, making these experiments widely available and reproducible across research labs. 
For more details, see S.\ref{app:extime}.

\section{Results}

\subsection{Tasks without shared structure -- Atari}

We chose a subset of Atari environments where the agent could achieve reasonable performance with less training and followed \cite{machado_revisiting_2018} in using sticky actions. 
The environments were presented with arbitrary ordering, as testing each permutation would have been prohibitively expensive.

\begin{figure}[tbh!]
  \centering
  \begin{subfigure}{0.48\textwidth}
    \includegraphics[width=\linewidth]{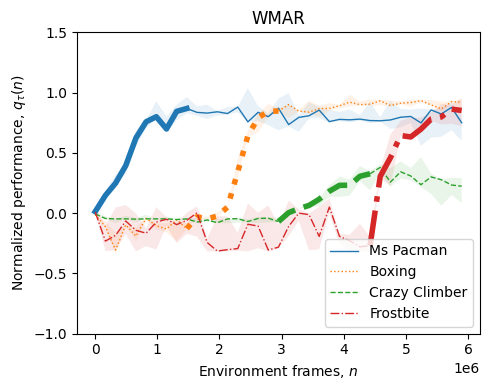}
   \end{subfigure}%
  \begin{subfigure}{0.48\textwidth}
    \includegraphics[width=\linewidth]{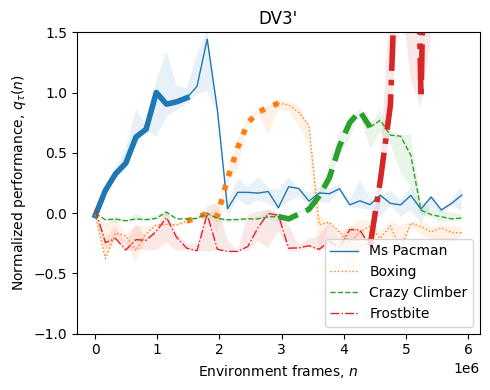}
  \end{subfigure}
  \caption{Tasks without shared structure. The line is the median and the shaded area is between the 0.25 and 0.75 quartiles, of 5 seeds. Bold line segments show when a task is being trained. Scores are normalised by eq.~\eqref{eq:normalization}.}
  \label{fig:atari-plot}
\end{figure}

Normalised performance is plotted in \autoref{fig:atari-plot}, and single-task results used for reward scaling are given in S.\ref{app:baseline}.
The predetermined reward scales are shown in S.Table~\ref{tab:atari_rew_scales}.
WMAR was capable of continual RL, whereas it is clear that \DV\ was not; it only performed well when it was actively training in a given task.

\paragraph{Forgetting (Backward transfer)} 
\DV\ struggled with retaining performance, losing nearly all knowledge after learning each new task. 
In contrast, WMAR maintained much of its performance in previously learnt tasks, demonstrating significantly improved stability. 
This highlighted the importance of the distribution matching buffer in preventing forgetting, especially when tasks differ greatly. 
Even though the probability of drawing samples from the first task decreased to 12.5\% by the end of training, WMAR retained much of its performance on the early tasks. However, this was not universal; for instance, `Crazy Climber' showed substantial forgetting and the worst overall performance, even with the buffer in place.

\paragraph{Forward transfer}
In general WMAR's ability to learn new tasks was maintained as it learnt new tasks, reaching close to single-task performance on all of the tasks.
There was one exception in the 3rd task, `Crazy Climber', which was approximately half as good as earlier tasks. 
On the other hand, \DV\ learnt slightly less effectively in each subsequent task.
However, \DV\ did not suffer on `Crazy Climber' and on the 4th and last task, `Frostbite', performance was well above single-task performance.
As a result, the average forward transfer for \DV\ was higher than for WMAR.

\begin{table}[tbh!]
    \caption{Average forgetting and forward transfer on tasks without shared structure (Atari). The metrics are shown as median with (0.25, 0.75) quartile confidence intervals, across 5 seeds, and calculated using normalised scores, eq.~\ref{eq:normalization}.}
    \label{tab:wmar_vs_wmar_fifo_atari}
    \centering
    \small
    \begin{tabular}{p{2cm} p{3.7cm} p{3.7cm}}
        \toprule
        \textbf{Model} & \textbf{Avg. forgetting} & \textbf{Avg. fwd. transfer} \\
                       & (Lower is better) & (Higher is better) \\
        \midrule
WMAR & $\mathbf{0.071}$ $(0.026,0.082)$ & $-0.392$ $(-0.432,-0.357)$ \\
\DV & $0.665$ $(0.636,0.810)$ & $\mathbf{1.049}$ $(0.245,1.246)$ \\
        \bottomrule
    \end{tabular}
\end{table}

\subsection{Tasks with shared structure -- CoinRun}

\begin{figure}[tb!]
  \centering
  \begin{subfigure}{0.49\textwidth}
	 \includegraphics[width=\linewidth]{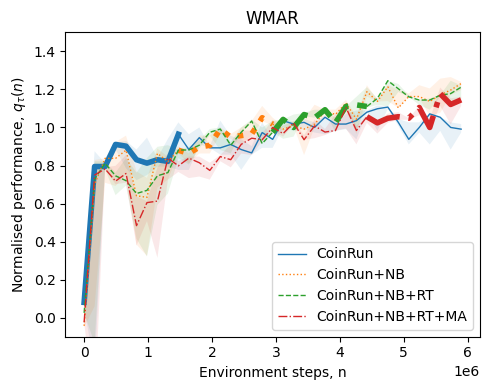}
   \end{subfigure}
   \hfill
  \begin{subfigure}{0.49\textwidth}
    \includegraphics[width=\linewidth]{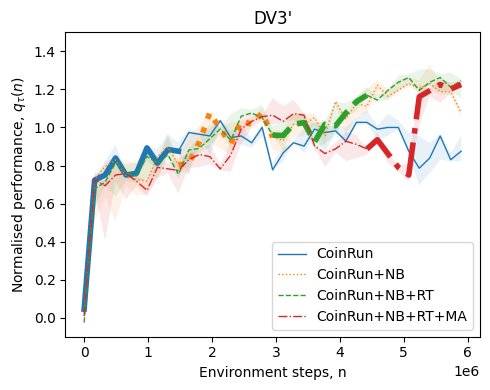}
  \end{subfigure}
  \caption{Tasks with shared structure. The line is the median and the shaded area is between the 0.25 and 0.75 quartiles, of 5 seeds. Bold line segments show when a task is being trained. Scores are normalised by eq.~\ref{eq:normalization}.}
  \label{fig:coinrun-plot}
\end{figure}

Normalised performance is shown in \autoref{fig:coinrun-plot}, and single-task results used for reward scaling are given in S.\ref{app:baseline}.
The terminology for CoinRun environments is: NB = no background, RT = restricted themes, and MA = monochrome assets.

Although CoinRun environments have identical mechanics, they share both subtle and substantial visual differences, posing significant challenges. 
However, WMAR could learn continuously and displayed forward and backward transfer of skills between environments.
\DV\ showed similar characteristics, but performance dropped on the first task, `CoinRun', after training on it (forgetting), and on the last task before training on it (lacking forward transfer).
They had approximately equal peak performance on each task; however, WMAR was noticeably more consistent than \DV.
Overall, WMAR showed more consistent performance across tasks with fewer fluctuations in performance during training, which may be a desirable property in practical applications.

\paragraph{Forgetting (Backward transfer)}
WMAR exhibited desirable forgetting (backward transfer), whereas \DV\ displayed very little backward transfer, see \autoref{tab:wmar_vs_wmar_fifo_coinrun}.

\paragraph{Forward transfer} 
Both WMAR and \DV\ had good forward transfer properties \autoref{tab:wmar_vs_wmar_fifo_coinrun}; the augmented buffer did not make a noticeable difference.
We found significant forward transfer on all tasks, greatly reducing the amount of time required to reach a given performance. 
It is notable as the tasks were constructed so that later task levels (aspects of a task) could be different from earlier tasks, making generalisation more challenging.

\begin{table}[thb!]
    \caption{Average forgetting and forward transfer on tasks with shared structure (Procgen CoinRun). The metrics are shown as median with (0.25, 0.75) quartile confidence intervals, across 5 seeds, and calculated using normalised scores, eq.~\ref{eq:normalization}.}
    \label{tab:wmar_vs_wmar_fifo_coinrun}
    \small
    \centering
    \begin{tabular}{p{2cm} p{3.7cm} p{3.7cm}}
        \toprule
        \textbf{Model} & \textbf{Avg. forgetting} & \textbf{Avg. fwd. transfer} \\
        & (Lower is better) & (Higher is better) \\
        \midrule
WMAR & $\mathbf{-0.077}$ $(-0.103,-0.062)$ & $\mathbf{0.316}$ $(0.286,0.357)$ \\
\DV & $-0.019$ $(-0.050,0.010)$ & $0.300$ $(0.239,0.335)$ \\
        \bottomrule
    \end{tabular}
\end{table}

\section{Discussion}

WMAR had superior performance with the same memory requirements compared to \DV.
On tasks without shared structure, WMAR exhibited almost no forgetting, while \DV\ had almost complete forgetting. 
On tasks with shared structure, WMAR showed noticeably decreased forgetting. 
In addition, WMAR was able to learn multiple tasks with highly varying magnitude of rewards, without task id's, which is an important and difficult challenge.
As RL algorithms often consume high compute, we emphasise the benefit of lower computational and memory costs.
Most RL is based on model-free approaches, where experience replay is used to improve learning the policy.
Our results confirm that replay to a world model in a model-based approach is also a strong foundation for continual RL.

Our methods are largely orthogonal to previous state-of-the-art approaches to combating catastrophic forgetting such as EWC and P\&C which work over network parameters, and CLEAR which uses replay but typically operates over model-free approaches and uses behaviour cloning of the policy from environment input to action output.

\paragraph{Generalisation}
Forward and backward transfer was far better for tasks with shared structure than those without shared structure. 
We hypothesise that this was due to the world model's convolutional feature extractor, which could benefit from commonality in the visual domain only. It may have suffered from more abstract changes such as inverting the colours or permuting the controls between tasks, leading to decreased overall performance.

\paragraph{Reward scaling without task id's}
While WMAR was sufficiently robust in learning multiple tasks without forgetting, even when reward scales are somewhat different, we found that this property did not hold when reward scales differed significantly (e.g. by a factor of $10^2$), where only the task with the higher reward scale would be learnt. 
In that case, we found that approximate reward scaling allowed the agent to learn multiple tasks.
We hypothesise that the different rewards and subsequent returns cause poorly scaled advantages when training the actor, resulting in the actor only learning tasks with the highest returns. 
Experiments with automatic scaling of advantages through non-linear squashing transformations proved to hurt learning on individual tasks, so the static, linear, reward transformation was used.

\paragraph{Memory capacity}

Despite the benefits and improved memory capacity of WMAR, a key limitation of any buffer-based method is finite capacity.
As more tasks are explored and previous tasks are not revisited, an increasing number of samples from previous tasks will inevitably be lost, leading to forgetting.


\subsection{Limitations and future work}

This study provides a strong justification for future work; however, the experiments were limited in scope. They could be expanded (e.g., as in \cite{hafner_dream_2020,kessler_effectiveness_2023}) by increasing the length of task sequences, increasing the number of environment steps, testing on more environments, and comparing to additional reference models.

Testing all task permutations is prohibitively expensive and therefore, we used a set task order. However, task ordering can result in significant differences in CL results (e.g., Appendix G in \cite{rahimi-kalahroudi_replay_2023}). 
Future work could randomise the order (as a proxy for testing all permutations) over a larger number of seeds.

Another area for future work is to improve WMAR by combining it with existing techniques such as behaviour cloning in CLEAR. Such an adaptation could counter shifts in the latent distribution as the world model trains and where the actor is frozen.

\section{Conclusions}
We extended a well-known model-based world model architecture, DreamerV3, with an augmented replay buffer, and applied it to the problem of continual RL in two scenarios: tasks with and without shared structure, i.e. commonalities between tasks that could be leveraged by an agent. 
WMAR and DV3' were set with the same memory budget (same-sized buffers) and were compared.
Performance was evaluated for forward and backward transfer, in addition to the common practice of measuring only forgetting.
We found that model-based agents are capable of continual learning on both task types.
The augmented replay buffer of WMAR conferred a minor benefit in tasks with shared structure and substantial improvements in tasks without shared structure.
The results suggest that model-based RL using a world model with a memory-efficient replay buffer can be an effective and practical approach to continual RL, justifying future work.

\ifdefined\mainsupp
\else

\bibliography{refs}
\bibliographystyle{splncs04}

\end{document}

\fi


\fi

\title{Supplementary Information for Augmenting Replay in World Models for Continual Reinforcement Learning}

\titlerunning{Supplementary Information}

\author{}
\authorrunning{}
%
\institute{}

\maketitle

\section{Background}
\label{app:background}

\subsection{Reinforcement learning (RL)}
\label{app:background_rl}

We explore the application of CL in finite, discrete time partially observable RL environments. Each environment may be characterised as a Partially Observable Markov Decision Process \cite{kaelbling_planning_1998} (POMDP), in which the agent does not have access to the state $s \in S$, and is a generalisation of the fully observable case \cite{puterman_chapter_1990}. A POMDP is a tuple $M=(\mathcal{S},\mathcal{A},p,r,\Omega,\mathcal{O},\gamma)$. $\mathcal{S}$ is the set of states, $\mathcal{A}$ is the set of actions. Where, at time step $t$, the current state and action $s_t\in\mathcal{S}$, $a_t\in\mathcal{A}$ and the next state $s_{t+1}$ are modelled by the stochastic state transition $s_{t+1} \sim p(s_t, a_t)$. The reward is given by $r:\mathcal{S}\times\mathcal{A}\times\mathcal{S}\to\mathbb{R}, r(s_t, a_t, s_{t+1})$. $\Omega$ is the set of observations, where the observation $\omega_t \in \Omega$ is modelled by $\omega_t \sim \mathcal{O}(s_t)$. The discount factor for computing returns is $\gamma \in (0,1)$.

Simultaneously, actions are sampled from a stochastic policy $a_t\sim\pi(\omega_t)$. The objective is to find a $\pi$ which maximises the expected discounted returns from any initial state $\mathbb{E}_\pi[R_0 \mid s_0]$, where under a finite horizon $T$, $R_t = \sum_{i=t}^T \gamma^{i-t}r(s_i,a_i,s_{i+1})$. Generally, the policy $\pi_\theta$ is a neural network parameterised by $\theta$. The policy may be learnt through a model-free approach, which means that no part of the learning system exists to simulate the environment. RL algorithms may also be classified as on- or off-policy methods. On-policy methods require that new samples from the environment be generated with the latest policy $\pi_\theta$ for each update. Off-policy methods must tackle the challenge of the discrepancy between the policy's current performance and its past performance when the sample was generated \cite{espeholt_impala_2018}. However, the ability to update the current policy using previously collected samples allows for better sample efficiency. An alternative is model-based approaches with learned world models. Here, the RL algorithm learns a simulator to model the state transitions $p$ and rewards $r$ using information from observations from past rollouts. Then, an RL algorithm may act in the simulator to learn the policy. Further, samples collected from any stage of learning may be stored in a replay buffer and used to train the simulator, making this method off-policy. This method involving the proxy of learning a simulator has also been shown to enable superior sample efficiency.

\subsection{Continual learning}

Although we exclusively explore the application and development of CL techniques in the RL setting, the problem of CL has also long existed in the supervised learning domain, attracting a wide body of research (e.g. \cite{kirkpatrick_overcoming_2017,lopez-paz_gradient_2017}). Neural networks have long been known to exhibit the catastrophic forgetting phenomenon. Indeed, the problem was first recognised within the context of supervised learning with neural networks \cite{mccloskey_catastrophic_1989,chen_continual_2018}, where it was discovered that training on new tasks or categories will likely cause a neural network to forget the learnings from previous tasks.

In the common supervised learning setting, a dataset may be continually shuffled and replayed, eliminating non-stationarity in the data distribution, ensuring balanced performance across all samples \cite{french_catastrophic_1999}. This approach has been shown to be an adequate solution to catastrophic forgetting. However, it may not be possible to sample a stationary data distribution in applications where large amounts of data are streamed, such as in RL where tasks are introduced sequentially and not reliably repeated.

\section{Related work}

In time, the challenges of CL were recognised and subsequently applied to RL. Here, the challenge of CL can be seen as a special case of a supervised learning task in off-policy RL.

Previous research has attempted to apply alternate training schemes supported by replay. However, these methods focus mainly on model-free RL algorithms \cite{isele_selective_2018,rolnick_experience_2019}. Generally, strategies to mitigate catastrophic forgetting can be grouped by rehearsal, regularisation, and parameter isolation methods. All of these methods work to reduce plasticity and forgetting and aim to stabilise learning; believing that it is stability, not plasticity, that is the key limiting factor. Rehearsal methods most commonly work by using replay of past data, whereas regularisation and parameter isolation methods work through constraining select parameters or the output of the neural network itself.

Synaptic consolidation approaches include regularisation and parameter isolation methods and aim to preserve the network parameters important to previously learned tasks. These approaches include Elastic Weight Consolidation (EWC) \cite{kirkpatrick_overcoming_2017} and Progress \& Compress (P\&C) \cite{schwarz_progress_2018}, which attempt to consolidate past experience at the level of individual neurons through constraining the movement of select weights. It is not obvious how experience replay may be used in synergy with those methods within an RL algorithm.

Other approaches focus on replaying experiences or generating typical experiences \cite{shin_continual_2017}, to improve agent training and avoid catastrophic forgetting. These methods have achieved strong results. The state-of-the-art CL algorithm in model-free RL is the CLEAR method \cite{rolnick_experience_2019}, which uses replay buffers augmented with V-trace importance sampling for off-policy learning and behaviour cloning to improve learning stability. Inspired in part by evidence from neuroscience, alternate augmentations to the replay buffer, such as selective experience replay, favouring surprise or reward, have also been proposed. Empirical data \cite{isele_selective_2018} from experiments examining the efficacy of various replay buffer augmentations, determined that matching the global training distribution to be an effective method of mitigating catastrophic forgetting without the requirement to store all experiences.

World models \cite{ha_world_2018} describe model-based RL approaches that are able to solve a variety of complex environments with superior sample efficiency \cite{hafner_mastering_2023,hafner_mastering_2022,schrittwieser_mastering_2020}. Research \cite{kessler_surprising_2022} related to the applicability of world models to continual learning has shown evidence of their potential in this domain. The work has also shown that the DreamerV2 \cite{hafner_mastering_2022} algorithm with a standard FIFO replay buffer that persists across tasks was able to curb catastrophic forgetting. Such an approach may be thought of as a memory approach. However, typical approaches to training world models and continual learning with world models demand large replay buffers, holding on the order of millions of samples, most notably in the form of images with a significant memory footprint. This poses a computational constraint and challenge to scalability. Rather, we are able to show that, through augmentation of the replay buffer, a significantly smaller, specifically crafted buffer can also curb catastrophic forgetting.

\section{World models}
\label{app:world_models}

\subsection{Training algorithm}
\label{app:training_algorithm}

\begin{algorithm}
\caption{WMAR training algorithm}
\label{alg:wmar}
\begin{algorithmic}
    \State \textbf{Hyperparameters: } World model training iterations $K$.
    \State \textbf{Input: } World model $M$, augmented replay buffer $\mathcal{D}$, sequence of tasks $\boldsymbol{\mathcal{T}}_{1:T} = (\tau_1, \tau_2, \dots, \tau_T)$.
    \For{$\tau = \tau_1, \tau_2, \dots, \tau_T$}
        \For{$i = 1, 2, \dots, K$}
            \State Train world model $M$ on $\mathcal{D}$.
            \State Train actor $\pi$ using $M$.
            \State Use $\pi$ in $\tau$ and append episodes to $\mathcal{D}$.
        \EndFor
    \EndFor
\end{algorithmic}
\end{algorithm}

\subsection{Network architecture}
\label{app:network_arch}

\begin{figure}
  \centering
  \includegraphics[width=0.6\textwidth]{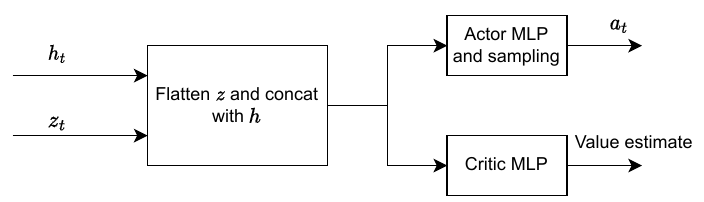}
  \caption{Actor critic definition.}
  \label{fig:actor_critic}
\end{figure}

The Actor Critic architecture is shown in Figure~\ref{fig:actor_critic}.
We adhered to most of the parameters and architectural choices of DreamerV3. Changes were primarily made to benefit wall time, as running continual learning experiments is computationally expensive. See \autoref{tab:hyperparameters}.

\begin{table*}
    \caption{Hyperparameters}
    \label{tab:hyperparameters}
    \centering
    \begin{tabular}{l l l l}
        \toprule
        Name & DreamerV3 & WMAR & \DV \\
        \midrule
        Replay capacity (FIFO) & 1M & 0.26M & 0.52M \\-
        Replay capacity (long-term) & 0 & 0.26M & 0 \\
        Batch size & 16 & 16 & 16 \\
        Batch length & 64 & 32 & 32 \\
        Activation & LayerNorm+SiLU & LayerNorm+SiLU & LayerNorm+SiLU \\
        Activation (GRU) & LayerNorm+SiLU & Tanh & Tanh \\
        \bottomrule
    \end{tabular}
\end{table*}

\paragraph{CNN encoder and decoder} Following DreamerV3, the convolutional feature extractor's input is a $64\times 64$ RGB image as a resized environment frame. The encoder convolutional neural network (CNN) \cite{lecun_backpropagation_1989} consists of stride 2 convolutions of doubling depth with the same padding until the image was at a resolution of $4\times 4$, where it is flattened. We elected to use the ``small'' configuration of the hyperparameters controlling network architecture from DreamerV3 to appropriately manage experiment wall time. Hence, 4 convolutional layers were used with depths of 32, 64, 128, and 256 respectively. As with DreamerV3 we used channel-wise layer normalisation \cite{ba_layer_2016} and SiLU \cite{hendrycks_gaussian_2023} activation for the CNN. The CNN decoder performs a linear transformation of the model state to a $4\times 4\times 256 = 4096$ vector before reshaping to a $4\times 4$ image and inverting the encoder architecture to reconstruct the original environment frame.

\paragraph{MLP} All multi-layer perceptrons (MLP) within the RSSM, actor, and critic are 2 layers and 512 hidden units in accordance with the ``small'' configuration of DreamerV3.

\subsection{Augmented replay buffer}

\begin{algorithm}
\caption{Sampling from combined buffers $\mathcal{D}_1$ and $\mathcal{D}_2$}\label{alg:combined_buffers}
	\begin{algorithmic}
		\State Combined (augmented) buffer $\mathcal{D} \dot{=} \{ \mathcal{D}_1, \mathcal{D}_2 \}$.
		\State Uniformly sample $i \in \{ 1, 2 \}$.
		\State \Return Sampled minibatch from $\mathcal{D}_i$.
	\end{algorithmic}
\end{algorithm}

%
%

\section{Experiments}

\subsection{Experimental parameters and execution time}
\label{app:extime}

Our method makes use of a number of environment samples significantly larger than the size of the replay buffer. Constraints on computational resources necessitate approximate conversions of GPU time. We follow \cite{hafner_mastering_2023} by estimating that the A40 is twice as fast as the V100. We also estimate that the A40 is marginally faster than the L4. The environment steps and samples, and the compute time in approximate GPU days are multiplied by the number of tasks in the CL suite.

\section{Results}
\label{app:results}

\subsection{Validation of DreamerV3 implementation}
\label{app:validation}

\begin{figure}[h!]
  \center
  \includegraphics[width=0.6\textwidth]{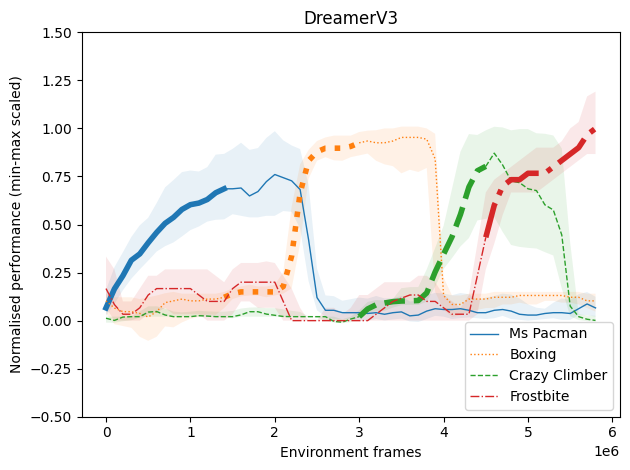}
  \caption{Performance of DreamerV3, with bold line segments denoting the periods in which certain tasks are being trained. Scores are normalised using min-max normalisation. The line is the median and the shaded area is between the 0.25 and 0.75 quantiles, of 5 seeds.}
  \label{fig:dv3}
\end{figure}

To validate that \DV\ was a faithful representation of DreamerV3, we compared \DV\ to an open source implementation by the author \url{https://github.com/danijar/dreamerv3}.
We tested it on the tasks without shared structure. 
See Figure~\ref{fig:dv3} and compare to Figure~\ref{fig:atari-plot}.

\subsection{Single-task runs}
\label{app:baseline}

The parameters used for single-tasks for CoinRun (shared structure) are shown in \autoref{tab:impl_comparison_coinrun} and for Atari (without shared structure) in \autoref{tab:impl_comparison_atari}.

\begin{table*}[h!]
    \caption{CoinRun single-task training parameters. GPU days are measured in A40 days.}
    \label{tab:impl_comparison_coinrun}
    \centering
    \begin{tabular}{l l l l l}
        \toprule
         & Env. steps & Env. samples & Approx. GPU days & Buffer size \\
        \midrule
        DreamerV3 & 1.6M & 1.6M & 0.25 & 1M \\
        WMAR & 1.5M & 1.5M & 0.25 & 0.26M $+$ 0.26M \\
        \DV & 1.5M & 1.5M & 0.25 & 0.52M \\
        \bottomrule
    \end{tabular}
\end{table*}

\begin{table*}[h!]
    \caption{Atari single-task training parameters. GPU days are measured in A40 days.}
    \label{tab:impl_comparison_atari}
    \centering
    \begin{tabular}{l l l l l}
        \toprule
         & Env. steps & Env. samples & Approx. GPU days & Buffer size \\
        \midrule
        DreamerV3 & 0.4M & 0.1M & 0.25 & 1M \\
        WMAR & 6M & 1.5M & 0.25 & 0.26M $+$ 0.26M \\
        \DV & 6M & 1.5M & 0.25 & 0.52M \\
        \bottomrule
    \end{tabular}
\end{table*}

The single-task results for Atari (no shared structure) are shown in \autoref{tab:results_atari} with the reward scales in \autoref{tab:atari_rew_scales}. 
The single-task results for CoinRun (shared structure) are shown in \autoref{tab:results_coinrun}.

\begin{table*}[h!]
    \caption{Atari single-task experimental results, median across 5 random seeds. Scores are unnormalised at the end of training.}
    \label{tab:results_atari}
    \centering
    \begin{tabular}{l l l l l}
        \toprule
        Task & Random & WMAR \\
        \midrule
        Ms Pacman     & 19.97 & 1661.25   \\
        Boxing        & 2.25 & 88.88     \\
        Crazy Climber & 7.79 & 100316.67 \\
        Frostbite     & 13.38 & 277.50    \\
        \bottomrule
    \end{tabular}
\end{table*}

\begin{table*}[h!]
    \caption{Atari environment reward scales for training WMAR.}
    \label{tab:atari_rew_scales}
    \centering
    \begin{tabular}{l l}
        \toprule
        Name & Reward scale \\
        \midrule
        Ms Pacman     & 0.05 \\
        Boxing        & 1 \\
        Crazy Climber & 0.001 \\
        Frostbite     & 0.2 \\
        \bottomrule
    \end{tabular}
\end{table*}

\begin{table*}[h!]
    \caption{Procgen CoinRun single-task experimental results, median across 5 random seeds. Scores are unnormalised at the end of training.}
    \label{tab:results_coinrun}
    \centering
    \begin{tabular}{l l l l l}
        \toprule
        Task & Random & WMAR \\
        \midrule
        CoinRun          & 2.30 & 6.68 \\
        CoinRun+NB       & 2.46 & 7.03 \\
        CoinRun+NB+RT    & 2.46 & 7.03 \\
        CoinRun+NB+RT+MA & 2.46 & 7.30 \\
        \bottomrule
    \end{tabular}
\end{table*}

\ifdefined\mainsupp
\else

\clearpage

\bibliography{refs}
\bibliographystyle{splncs04}